\documentclass{article} 
\usepackage[preprint]{colm2026_conference}

\usepackage{microtype}
\usepackage{hyperref}
\usepackage{url}
\usepackage{booktabs}
\usepackage{booktabs}
\usepackage{tabularx}
\usepackage{array}
\usepackage{graphicx}
\usepackage{subcaption}
\usepackage{pifont}
\usepackage{amsmath}

\usepackage{lineno}

\definecolor{darkblue}{rgb}{0, 0, 0.5}
\hypersetup{colorlinks=true, citecolor=darkblue, linkcolor=darkblue, urlcolor=darkblue}

\title{A Modular Architecture for \\ Typologically Controlled Lexicon Generation}

\author{Sankalp Tattwadarshi Swain, Dhruv Kumar \\
Birla Institute of Technology and Science, Pilani \\
Pilani, Rajasthan 333031, India \\
\texttt{\{f20230769, dhruv.kumar\}@pilani.bits-pilani.ac.in}
}

\begin{document}

\ifcolmsubmission
\linenumbers
\fi

\maketitle

\begin{abstract}
Constructing artificial lexicons that are pronounceable,
typologically plausible, and semantically structured remains
an open challenge in computational linguistics.
Existing conlang generators either lack formal phonotactic
guarantees or delegate generation to opaque, non-reproducible
LLM-based pipelines.
We propose a modular framework that samples phoneme inventories
from PHOIBLE, generates word forms under interchangeable phonological
grammars (deterministic, OT, and MaxEnt), and
assigns meanings via a Swadesh--Leipzig--Jakarta ontology with
explicit form--meaning alignment.
Evaluation on character $n$-gram perplexity, log-likelihood,
and KL divergence against PHOIBLE across lexicon sizes of
100-5,000 forms shows that probabilistic grammars consistently
outperform deterministic and random baselines on both phonotactic
coherence and typological realism.
\end{abstract}

\section{Introduction}
\label{sec:introduction}
 
Modelling phonological and lexical structure remains central to
understanding how language systems can be formally represented,
learned, and generated \citep{prince2004optimality,hayes2008maximum}.
A practically important instance of this problem is
\emph{typologically grounded lexicon generation}: constructing
artificial lexicons that map pronounceable, phonotactically
well-formed word forms to meanings while remaining aligned with
cross-linguistic phonological patterns.
Existing approaches are either rule-based tools that offer
designer control but lack formal phonotactic guarantees and
typological grounding \citep{heyer2021conlangs,cai2023procedural},
or LLM-based pipelines that produce plausible output but surrender
interpretability and reproducibility entirely
\citep{alper2025conlangcrafter,taguchi2025iasc}.
Critically, no existing system enables a researcher to hold the
phoneme inventory and constraint set fixed while varying the
phonological grammar — the minimum requirement for isolating the
contribution of any single formalisation to lexical structure.
 
We introduce a modular, fully parameterized lexicon generation
framework that resolves this gap by decomposing the pipeline into
three independently controllable components.
First, a phonological inventory sampler draws phoneme inventories
from PHOIBLE frequency distributions \citep{moran2019phoible},
enforcing empirically validated implicational universals through
a statistically grounded repair stage.
Second, a word generation module evaluates candidate forms
against a shared constraint set under four interchangeable
grammatical formalisms — deterministic, Optimality Theory (OT),
Harmonic Grammar (HG), and Maximum Entropy (MaxEnt) — that
differ only in how constraint violations are resolved, ensuring
that every generated word form is pronounceable by construction.
Third, a semantic assignment module maps generated forms to
meanings drawn from a merged Leipzig--Jakarta and Swadesh
ontology, optimising a Spearman rank correlation between
phonological and semantic distance to produce a lexicon in
which form--meaning alignment is an explicit objective rather
than an incidental property.
 
We evaluate the framework quantitatively along two complementary
axes across lexicon sizes from 100 to 5,000 word forms.
Results show that probabilistic grammars — OT and MaxEnt —
consistently outperform the deterministic and random baselines
on both axes, and that the deterministic grammar pays a
measurable typological cost through categorical constraint
enforcement that systematically skews the phoneme distribution.This work is relevant to the language modelling community as both
a controlled testbed for probing phonotactic knowledge in LLMs and
a pipeline for generating pronounceable, typologically grounded
synthetic corpora for low-resource speech and language modelling.
 
\paragraph{Contributions.} Our main contributions are as follows:
 
\begin{itemize}
 
  \item \textbf{Modular, parameterized lexicon generation.}
We propose a unified framework in which deterministic, OT, HG,
and MaxEnt phonological grammars operate over a shared inventory
and shared constraint set. The novelty lies not in the grammars
themselves but in their integration with typologically grounded
inventory sampling, enabling confound-free empirical comparison
across formalisms that existing systems do not support.
 
  \item \textbf{Typologically grounded inventory sampling.}
  We integrate PHOIBLE frequency distributions and statistically
  validated implicational universals as structured generative
  priors for phoneme inventory construction, grounding generated
  lexicons in cross-linguistic typological reality.
 
  \item \textbf{Pronounceability as a formal guarantee.}
  By enforcing phonotactic constraints — including the Sonority
  Sequencing Principle, onset/coda structure, and nasal--stop
  homorganicity — at generation time, the framework guarantees
  that every output word form is well-formed, making the lexicon
  a principled foundation for spoken constructed languages.
 
  \item \textbf{Form--meaning alignment via semantic optimisation.}
  We introduce a hill-climbing semantic assignment procedure that
  maximises Spearman rank correlation between phonological and
  ontological semantic distance, operationalising lexical
  iconicity as an explicit generation objective.
 
  \item \textbf{Quantitative evaluation framework.}
  We introduce a joint evaluation protocol combining
  character-level language model perplexity and average
  log-likelihood with KL divergence against a typological
  reference distribution, providing a quantitative
  benchmark for comparing phonotactic well-formedness and
  typological realism of \emph{generated} lexicons.

  \item \textbf{Cross-grammar compatibility analysis.} We empirically
demonstrate that OT and MaxEnt induce distributionally equivalent
phonotactic spaces, and that deterministic grammars generate a strict
subset of probabilistic phonotactic patterns.
 
\end{itemize}

\begin{table*}[t]
\centering
\footnotesize
\setlength{\tabcolsep}{3pt}
\renewcommand{\arraystretch}{1.3}
\begin{tabular}{
  >{\raggedright\arraybackslash}p{2.5cm}
  >{\raggedright\arraybackslash}p{2.0cm}
  >{\raggedright\arraybackslash}p{2.1cm}
  >{\raggedright\arraybackslash}p{2.2cm}
  >{\raggedright\arraybackslash}p{1.7cm}
  >{\raggedright\arraybackslash}p{2.2cm}
}
\toprule
\textbf{System}
  & \textbf{Inventory Source}
  & \textbf{Word Form Generation}
  & \textbf{Phonotactics}
  & \textbf{Semantics}
  & \textbf{Typological Grounding} \\
\midrule
 
\citet{heyer2021conlangs}
  & User-defined templates
  & Template sampling
  & Hard rule filters (user-specified)
  & None
  & None \\
 
\citet{cai2023procedural}
  & Designer-specified parameters
  & Syllable template expansion
  & Configurable rules; no formal grammar
  & None
  & None \\
 
\citet{goldwater2003learning}
  & Real language phoneme sets
  & MaxEnt-ranked OT candidates
  & MaxEnt over ranked OT constraints
  & None
  & None \\
 
\citet{hayes2008maximum}
  & Real phoneme inventories
  & MaxEnt-weighted candidate selection
  & Maximum Entropy with learned constraint weights
  & None
  & None \\
 
\citet{futrell2017generative}
  & Real lexicons (14 languages)
  & Bayesian generative model (stochastic memoization)
  & Probabilistic; phonological feature hierarchy
  & None
  & Multi-lingual; no typological sampling \\
 
\citet{pimentel2020phonotactic}
  & Real lexicons (106 languages)
  & Not generative; complexity measurement only
  & LSTM / trigram LM; bits-per-phoneme measure
  & None
  & Cross-linguistic comparison; no inventory sampling \\
 
\citet{alper2025conlangcrafter}
  & LLM-generated phonology sketch
  & LLM lexicon generation
  & Implicit in LLM; not formally specified
  & LLM-generated glosses \& translations
  & WALS features (macro-level diversity only) \\
 
\citet{taguchi2025iasc}
  & LLM-generated via agentic refinement
  & Morphemes from annotated corpus
  & LLM-generated executable grammar
  & Corpus-derived from English glosses
  & None \\
 
\midrule
 
\textbf{This work}
  & \textbf{PHOIBLE frequency-weighted sampling}
  & \textbf{Formal grammar over sampled inventory (Det / OT / HG / MaxEnt)}
  & \textbf{Explicit constraint set; four interchangeable formalisms}
  & \textbf{Leipzig--Jakarta \& Swadesh ontology}
  & \textbf{PHOIBLE implicational universals; KL vs.\ global phoneme distribution} \\
 
\bottomrule
\end{tabular}
\caption{Comparison of lexicon generation systems across five dimensions:
inventory source, word form generation, phonotactics, semantics, and
typological grounding. Our system is the only one that unifies all five
within a single pipeline.}
\label{tab:comparison}
\end{table*}

\section{Background}
\label{sec:background}
 
\paragraph{Formal phonological grammars.}
Optimality Theory formalises phonotactic well-formedness through a
ranked constraint hierarchy under strict domination
\citep{prince2004optimality}, while Harmonic Grammar relaxes this
into a numeric weighting scheme \citep{smolensky2006harmonic},
and Maximum Entropy grammar extends it further into a fully
probabilistic model over weighted constraint violations
\citep{goldwater2003learning,hayes2008maximum}.
These frameworks have each been studied and applied to individual
natural languages in isolation.
However, a unified architecture in which deterministic, OT, HG,
and MaxEnt models operate over a shared inventory and shared
constraint set — enabling direct, confound-free empirical
comparison — has not been previously proposed.
 
\paragraph{Typological databases and phoneme inventory modelling.}
Cross-linguistic resources such as PHOIBLE \citep{moran2019phoible}
aggregate phoneme inventories from thousands of languages, encoding
implicational universals and quantitative markedness distributions
that demonstrate phoneme inventories are structured objects shaped
by universal typological pressures.
Despite the richness of this data, the integration of typological
databases as structured generative priors for phoneme inventory
sampling within a language generation pipeline has not been
addressed.
 
\paragraph{Rule-based and procedural conlang generation.}
\citet{heyer2021conlangs} and \citet{cai2023procedural} demonstrate
that automated conlang construction is feasible through
user-defined phoneme templates and parameterisable syllable
pipelines.
However, neither system grounds its inventory in cross-linguistic
data, employs a formal phonological grammar, or provides any
quantitative measure of the phonotactic well-formedness or
typological plausibility of the generated output — a gap that
prevents rigorous evaluation or reproducible comparison across
parameter settings.
 
\paragraph{Neural and probabilistic phonotactic models.}
\citet{futrell2017generative} introduce a Bayesian generative model
that captures sub-lexical phonotactic structure across typologically
diverse languages, and \citet{pimentel2020phonotactic} establish a
neural bits-per-phoneme measure of phonotactic complexity across
106 languages using LSTM language models.
While these contributions demonstrate that probabilistic and neural
methods can characterise the phonotactic structure of natural
lexicons with considerable precision, the application of such
models to the \emph{generation} of new lexicons under controlled
typological conditions and with structured semantic assignment
has not been addressed.
 
\paragraph{Large language model pipelines for conlang generation.}
\citet{alper2025conlangcrafter} introduce ConlangCrafter, a multi-hop
LLM pipeline evaluated through macro-level WALS-feature diversity
metrics, and \citet{taguchi2025iasc} present IASC, an agentic system
generating phonotactic grammars as executable code. Both delegate
phonotactic decisions to an LLM, producing output that is opaque,
non-reproducible, and without formal well-formedness guarantees.
Neither evaluates phonotactic quality at the word level nor supports
controlled ablation over grammatical formalisms.

\section{Methodology}
\label{sec:methodology}
 
\subsection{Overview}
\label{sec:overview}
 
Our lexicon generation framework comprises two principal components:
(1) a statistically grounded phonological inventory sampler, and
(2) a modular word generation system that operates over the sampled
inventory. The pipeline first constructs a typologically plausible
phoneme inventory by drawing on cross-linguistic frequency
distributions from PHOIBLE \citep{moran2019phoible}. This inventory
then serves as the fixed, shared foundation for four interchangeable
phonotactic grammars—deterministic, Optimality Theory (OT), Harmonic
Grammar (HG), and Maximum Entropy (MaxEnt)—which share the same
constraint representations but differ fundamentally in how violations
are evaluated and resolved. This design deliberately separates
typological realism, modelled at the inventory level, from
phonotactic variation, modelled at the grammatical level, enabling
controlled comparison across categorical and probabilistic
phonological formalisms within a single unified architecture.
 
\subsection{Phonological Inventory Sampling}
\label{sec:inventory}
 
The phonological sampler models sound systems as samples drawn from
an empirically structured typological distribution. Each phoneme is
assigned a base sampling probability proportional to its cross-linguistic
frequency across PHOIBLE inventories, thereby operationalising
markedness in probabilistic terms: typologically common segments are
preferentially selected over rare ones, without categorical exclusion
of any attested segment type.
 
Inventory sizes are constrained by cross-linguistic distributions.
Consonant and vowel counts are sampled such that the
vowel-to-consonant ratio falls within the range $[0.15,\, 0.40]$,
reflecting attested typological ranges \citep{moran2019phoible}.
Optional archetype modes restrict sampling to particular structural
subspaces—for instance, small CV-dominant systems or consonant-rich
systems—enabling controlled typological variation while preserving
empirical plausibility.
 
To capture implicational universals and co-occurrence tendencies, we
conduct feature-level statistical analysis over PHOIBLE inventories.
To validate the statistical basis for these typological constraints,
we computed Pearson correlation coefficients and chi-square statistics
over all PHOIBLE inventories (Table~\ref{tab:corr_chisq}). Correlation
analysis reveals that inventory size is positively associated with the
presence of marked segment classes: consonant count correlates
significantly with ejective presence ($r = 0.324$, $p < 0.001$) and
click presence ($r = 0.188$, $p < 0.001$), while vowel count
correlates with vowel length contrast ($r = 0.412$, $p < 0.001$),
justifying the conditioning of marked segment sampling probabilities
on inventory size. For several feature pairs, chi-square tests
yielded $\chi^2 = 0$ with $p = 1.0$, a result that initially suggests
independence but upon inspection reflects near-deterministic
implicational structure with empty or near-empty contingency cells.
Because chi-square assumes symmetric variation, it is an inappropriate
test for directional universals; we therefore turn to conditional
probabilities $P(Y \mid X)$ as a direct measure of implicational
strength (Table~\ref{tab:cond_prob}). This analysis reveals that four
of the five incorporated implicational universals are fully categorical
($P = 1.0$), while the remaining one is a near-universal ($P =
0.997$, failure rate $= 0.003$). Collectively, these results confirm
that the constraints encoded in the sampler reflect statistically
validated typological regularities rather than heuristic assumptions,
providing an empirical grounding for both the deterministic enforcement
of true universals and the probabilistic enforcement of near-universals.

\begin{table}[t]
\centering
\small
\begin{tabular}{llcc}
\toprule
\textbf{Relationship} & \textbf{Statistic} & \textbf{Value} & \textbf{\textit{p}-value} \\
\midrule
Consonant count vs.\ ejective presence  & $r$ & $0.324$ & $<0.001$ \\
Consonant count vs.\ click presence     & $r$ & $0.188$ & $<0.001$ \\
Inventory size vs.\ click presence      & $r$ & $0.188$ & $<0.001$ \\
Vowel count vs.\ vowel length contrast  & $r$ & $0.412$ & $<0.001$ \\
\midrule
Ejective $\leftrightarrow$ uvular               & $\chi^2$ & $0.000$ & $1.000$ \\
Pharyngeal $\leftrightarrow$ uvular             & $\chi^2$ & $0.000$ & $1.000$ \\
Nasal vowel $\leftrightarrow$ oral vowel        & $\chi^2$ & $0.000$ & $1.000$ \\
Voiced $\leftrightarrow$ voiceless obstruent    & $\chi^2$ & $0.656$ & $0.418$ \\
Fricative $\leftrightarrow$ stop                & $\chi^2$ & $0.000$ & $1.000$ \\
Front rounded $\leftrightarrow$ front unrounded & $\chi^2$ & $0.000$ & $1.000$ \\
\bottomrule
\end{tabular}
\caption{Pearson correlation coefficients ($r$) and chi-square statistics ($\chi^2$)
computed over PHOIBLE inventories. Correlation values measure symmetric associations
between inventory size and marked segment presence; chi-square values reflect
feature co-occurrence tests. All correlation results are significant at $p < 0.001$.}
\label{tab:corr_chisq}
\end{table}
 
\begin{table}[t]
\centering
\small
\begin{tabular}{lrrrr}
\toprule
\textbf{Implication} & \textbf{Total $X$} & \textbf{Violations} & $P(Y \mid X)$ & \textbf{Failure Rate} \\
\midrule
Pharyngeal $\Rightarrow$ uvular            & $15$   & $0$ & $1.000$ & $0.00000$ \\
Nasal vowel $\Rightarrow$ oral vowel       & $688$  & $0$ & $1.000$ & $0.00000$ \\
Voiced obs.\ $\Rightarrow$ voiceless obs.\ & $2355$ & $6$ & $0.997$ & $0.00255$ \\
Fricative $\Rightarrow$ stop               & $3020$ & $0$ & $1.000$ & $0.00000$ \\
Front rounded $\Rightarrow$ front unrounded & $273$ & $0$ & $1.000$ & $0.00000$ \\
\bottomrule
\end{tabular}
\caption{Conditional probability results for the five implicational
universals incorporated into the sampler. \emph{Total $X$} denotes
the number of inventories in which the antecedent feature is present;
\emph{Violations} counts cases where the consequent feature is absent;
$P(Y \mid X)$ is the empirical conditional probability; and
\emph{Failure Rate} is the complementary proportion. Four of the five
implications are fully categorical ($P = 1.0$); the voiced--voiceless
obstruent implication is a near-universal ($P = 0.997$).}
\label{tab:cond_prob}
\end{table}
 
\subsection{Word Generation Framework}
\label{sec:wordgen}
 
All word generators operate over the same sampled phoneme inventory
and share a unified constraint set encoding widely attested
phonotactic generalisations: \textsc{Onset}, \textsc{NoCoda},
\textsc{*Complex}, the Sonority Sequencing Principle (SSP), and
nasal-stop homorganicity. Candidate word forms are produced by
sampling syllable structures from a parameterised template and
filling syllable positions with segments drawn from the inventory.

\begin{figure}[t]
    \centering
    \includegraphics[width=\linewidth]{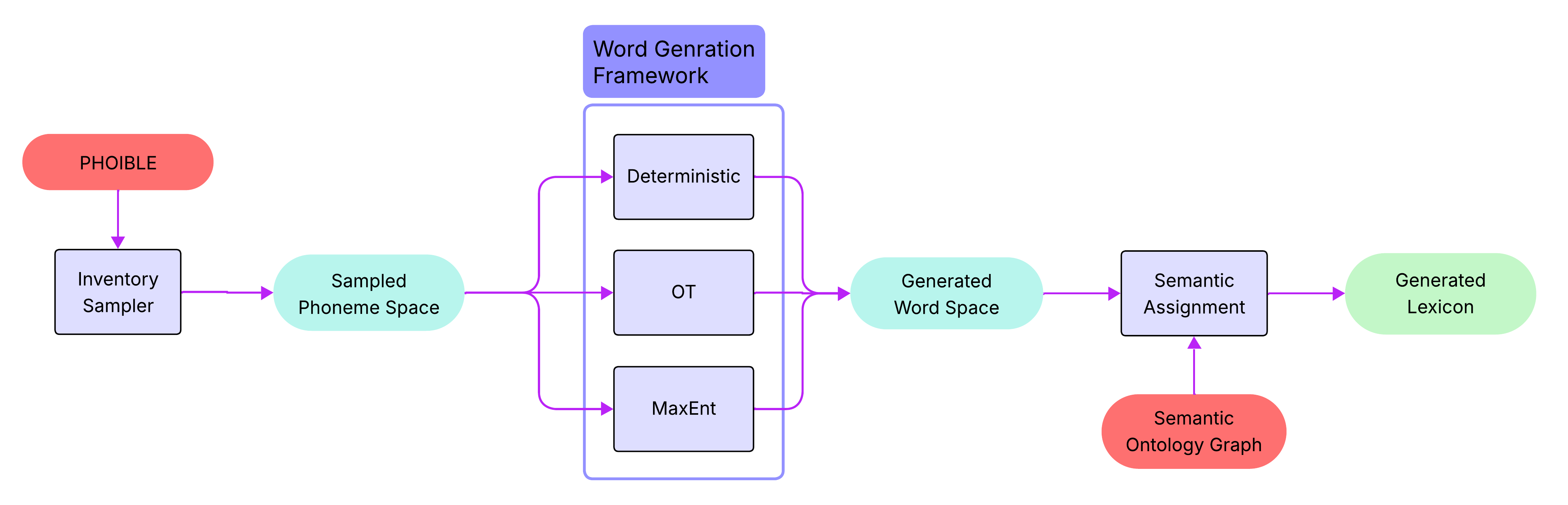}
    \caption{Overview of the lexicon generation pipeline. PHOIBLE-sampled inventories feed four interchangeable phonological grammars; the generated word space is then mapped to meanings via the semantic assignment module.}
    \label{fig:pipeline}
\end{figure}
 
\subsubsection{Deterministic Generator}
\label{sec:deterministic}
 
The deterministic phonotactic generator implements a rule-based
grammar in which constraints function as hard filters. Candidate
forms that violate structural limits (e.g., maximum onset
complexity), the SSP as formalised under \citep{clements1990role}
sonority hierarchy, or the homorganicity requirement for nasal–stop
sequences are rejected outright. This corresponds to the classical
generative view in which phonotactic well-formedness is binary and
constraint violations are categorically illicit
\citep{chomsky1968sound}.
 
\subsubsection{Optimality Theory Generator}
\label{sec:ot}

The Optimality-Theoretic (OT) generator replaces categorical
rejection with evaluation over a fully generated candidate set.
Three variants are implemented within a shared constraint
architecture: \textbf{Strict OT}, in which constraints are
hierarchically ranked under strict domination such that any
higher-ranked violation is fatal \citep{prince2004optimality};
\textbf{Stochastic OT}, which preserves this ranking but adds
Gaussian noise at evaluation time to yield probabilistic variation
\citep{boersma1997learning}; and \textbf{Harmonic Grammar (HG)},
which replaces strict ranking with numeric weights, computing
harmony as a weighted sum of violations and permitting multiple
lower-weighted violations to collectively outweigh a single
higher-weighted one \citep{smolensky2006harmonic,legendre1990harmonic}.
Although all three variants share the same constraint definitions,
they differ fundamentally in how constraint interaction is
formalised.
 
\subsubsection{Maximum Entropy Generator}
\label{sec:maxent}
 
The Maximum Entropy (MaxEnt) grammar extends Harmonic Grammar into a
fully probabilistic model. Harmony is computed as a weighted sum of
violations, and candidate probabilities are derived via the softmax
transformation $P(w) \propto e^{-H(w)}$ \citep{goldwater2003learning,
hayes2008maximum}, yielding a proper probability distribution over
surface forms. Unlike winner-take-all OT, MaxEnt assigns gradient
acceptability to candidates and supports intrinsic variation, making
it particularly well-suited for modelling probabilistic phonotactics
and lexical frequency distributions.

\subsection{Semantic Ontology and Meaning Space}
\label{sec:ontology}
 
We construct a hierarchical ontology graph in which each node
carries a canonical identifier and a human-readable gloss, with
parent--child edges encoding hyponymic relations. The meaning space
is derived by merging the Leipzig--Jakarta and Swadesh lists,
producing a typologically motivated core vocabulary. Lexicalization
operates exclusively at leaf nodes, preventing semantic collapsing.
Leaf concepts are sampled to match the target lexicon size, weighted
by pre-leaf parent category to ensure breadth across the ontology.
 
\subsection{Semantic Assignment}
\label{sec:semantic_assignment}
 
Given $N$ selected meanings $\mathcal{M} = \{m_1, \ldots, m_N\}$
and $N$ generated word forms $\mathcal{W} = \{w_1, \ldots, w_N\}$,
the module seeks a bijection $\sigma : \mathcal{M} \rightarrow
\mathcal{W}$ such that phonological similarity tracks semantic
similarity — a form--meaning alignment objective grounded in lexical
iconicity \citep{dingemanse2015arbitrariness}.
 
\paragraph{Semantic distance.}
Distance between meanings is computed as tree path distance over
the ontology:
\begin{equation}
    d_{\mathrm{sem}}(m_i, m_j) =
    \mathrm{depth}(m_i) + \mathrm{depth}(m_j)
    - 2 \cdot \mathrm{depth}\!\left(\mathrm{lca}(m_i, m_j)\right),
    \label{eq:sem_dist}
\end{equation}
where $\mathrm{lca}(m_i, m_j)$ denotes their lowest common ancestor.
 
\paragraph{Form distance.}
Phonological distance is measured using Levenshtein edit distance
$d_{\mathrm{form}}(w_i, w_j)$, a language-agnostic and
computationally tractable proxy for phonological similarity.
 
\paragraph{Alignment objective and optimisation.}
Mapping quality is assessed by the Spearman rank correlation
between semantic and form distance vectors over all meaning pairs:
\begin{equation}
    \mathcal{S}(\sigma) =
    \rho_s\!\left(
        \bigl[d_{\mathrm{sem}}(m_i, m_j)\bigr]_{i < j},\;
        \bigl[d_{\mathrm{form}}(\sigma(m_i), \sigma(m_j))\bigr]_{i < j}
    \right).
    \label{eq:spearman}
\end{equation}
Since exhaustive search over all $N!$ bijections is intractable,
we employ hill-climbing with random restarts: each restart proposes
random pairwise swaps $\sigma(m_i) \leftrightarrow \sigma(m_j)$,
retaining only improvements to $\mathcal{S}$. The dominant cost is
$O(N^2)$ per iteration, tractable for the lexicon sizes considered.
The final lexicon is stored as a JSON mapping from canonical
semantic identifiers to surface word forms.

\section{Evaluation}
\label{sec:evaluation}
 
\subsection{Phonotactic Well-Formedness}
\label{sec:eval_phonotactic}
 
Phonotactic well-formedness is evaluated using a smoothed character-level
$n$-gram language model trained on each generated lexicon. For a lexicon
of size $N$, a model is trained on the word forms produced by the
corresponding generator and evaluated on a held-out sample. We report
two metrics: \emph{perplexity} (lower is better) and \emph{average
log-likelihood} (higher, i.e., less negative, is better). Both metrics
are computed as a function of lexicon size, ranging from $100$ to
$5000$ word forms, to assess how structural coherence scales with
vocabulary size. In addition, we report an improvement ratio for each
constrained model relative to the random baseline, providing a
normalised measure of the benefit conferred by grammatical constraint
enforcement.
 
\subsection{Typological Realism}
\label{sec:eval_typological}
 
Typological realism is evaluated by measuring the divergence between
the phoneme frequency distribution of a generated lexicon and the
global phoneme frequency distribution derived from PHOIBLE
\citep{moran2019phoible}. Concretely, each generated lexicon is
tokenised into its constituent phonemes, and the resulting unigram
distribution $P$ is compared against the PHOIBLE reference
distribution $Q$ using \emph{Kullback--Leibler (KL) divergence}
$D_{\mathrm{KL}}(P \,\|\, Q)$ (lower is better). KL divergence is computed across lexicon sizes
from $100$ to $5000$, providing a trajectory of typological alignment
as vocabulary size grows.

\subsection{Cross-Grammar Compatibility}
\label{sec:eval_cross}
 
To assess the distributional relationships between phonological
formalisms, we construct a cross-grammar evaluation matrix. For
each grammar $g \in \{\text{deterministic, OT, MaxEnt, random}\}$,
a character-level neural language model is trained on a lexicon
generated by $g$ and evaluated on lexicons generated by every other
grammar. This yields a $4 \times 4$ matrix of average log-likelihood
and perplexity values, where entry $(g_{\text{train}},
g_{\text{test}})$ reflects how well the phonotactic distribution
learned from $g_{\text{train}}$ explains word forms generated by
$g_{\text{test}}$. Diagonal entries capture within-grammar fit;
off-diagonal entries reveal cross-grammar transferability. All
models are trained and evaluated at a fixed lexicon size of
$N = 5000$.
 
\section{Results}
\label{sec:results}
 
\subsection{Phonotactic Well-Formedness}
\label{sec:results_phonotactic}
 
\begin{figure}[t]
    \centering
    \begin{subfigure}{0.48\linewidth}
        \centering
        \includegraphics[width=\linewidth]{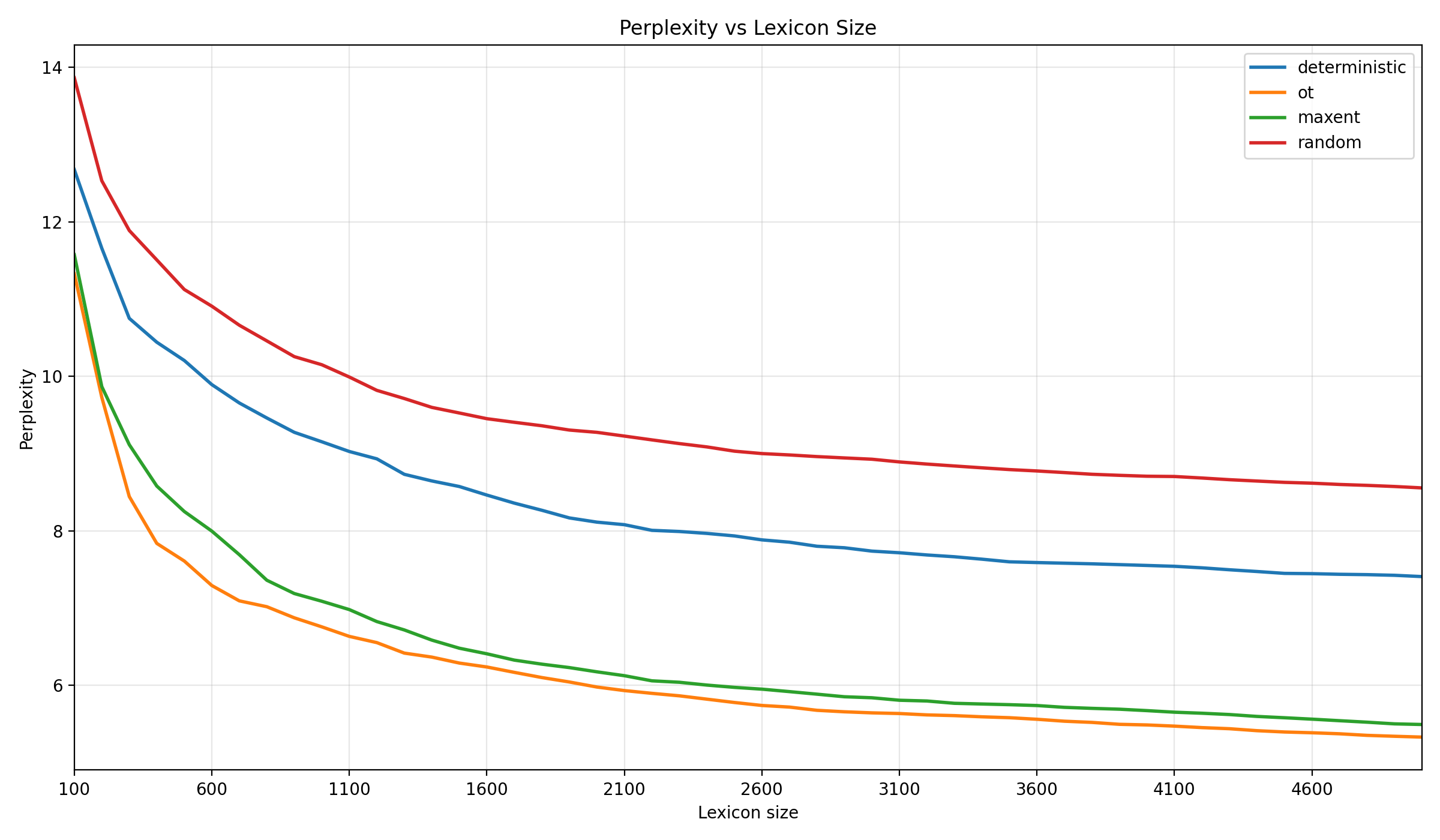}
        \caption{Character-level perplexity (lower is better).}
        \label{fig:perplexity}
    \end{subfigure}
    \hfill
    \begin{subfigure}{0.48\linewidth}
        \centering
        \includegraphics[width=\linewidth]{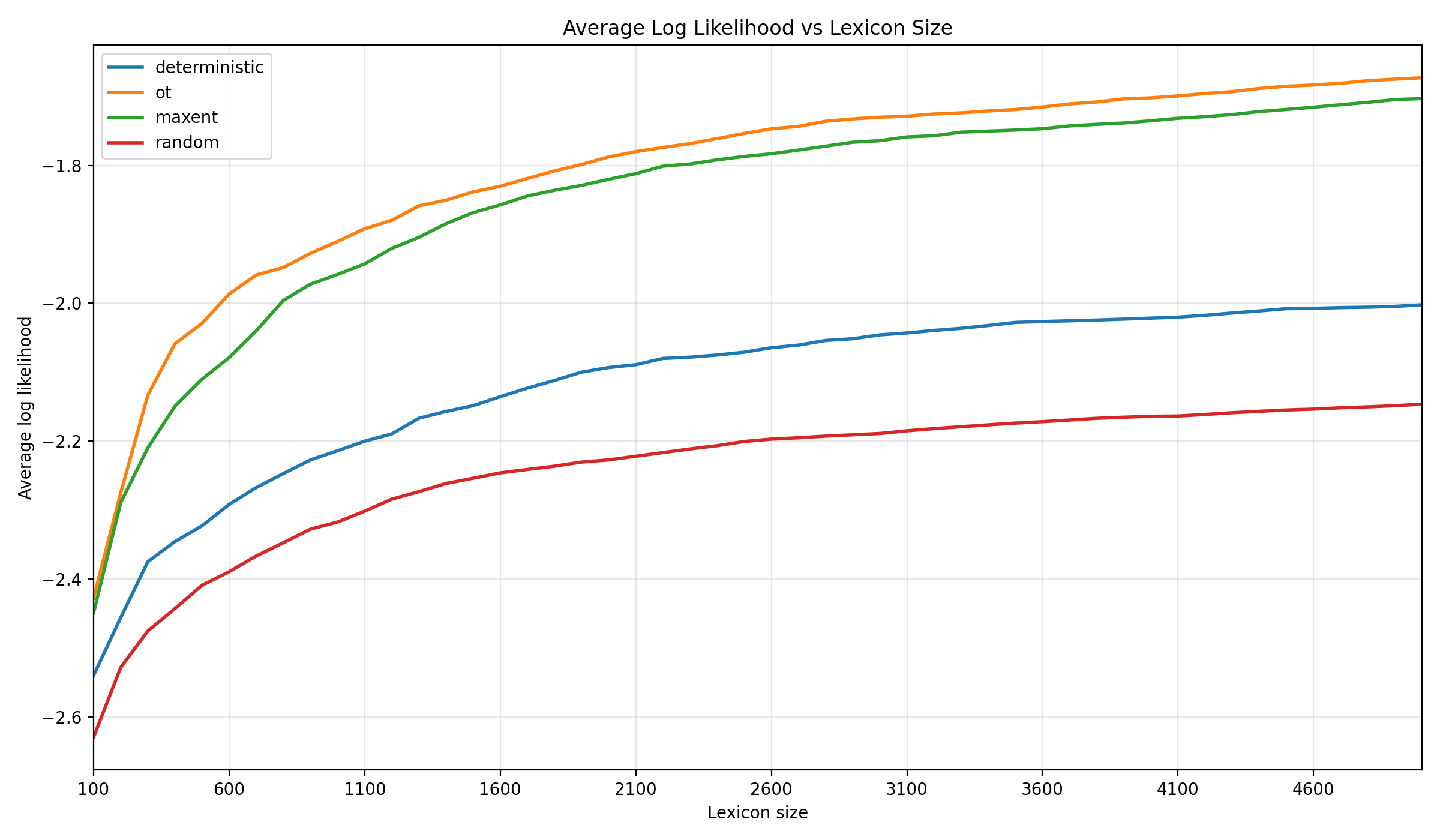}
        \caption{Average log-likelihood (higher is better).}
        \label{fig:loglik}
    \end{subfigure}
    \caption{Phonotactic well-formedness as a function of lexicon size.
    OT and MaxEnt consistently outperform the deterministic and random
    baselines across all lexicon sizes on both metrics.}
    \label{fig:phonotactic}
\end{figure}
 
Figure~\ref{fig:phonotactic} reveals that probabilistic constraint
evaluation produces measurably more learnable phonotactic structure
than categorical filtering. OT and MaxEnt converge to substantially
lower perplexity (${\approx}5.2$ and $5.4$ respectively at $N=5000$)
than the deterministic generator (${\approx}7.4$), which in turn
outperforms the random baseline (${\approx}8.6$). The consistent
ranking across both metrics — and its stability over the full
$100$--$5000$ size range — indicates that this is a property of the
grammars themselves rather than a sample-size artifact. The close
alignment of OT and MaxEnt suggests that probabilistic constraint
weighting, whether winner-take-all or distributional, produces
similarly coherent phonotactic regularities; the deterministic
generator's gap reflects the cost of hard-filtering, which eliminates
consistent structural patterns rather than penalising them
probabilistically.
 
\subsection{Typological Realism}
\label{sec:results_typological}
 
\begin{figure}[t]
    \centering
    \includegraphics[width=0.75\linewidth]{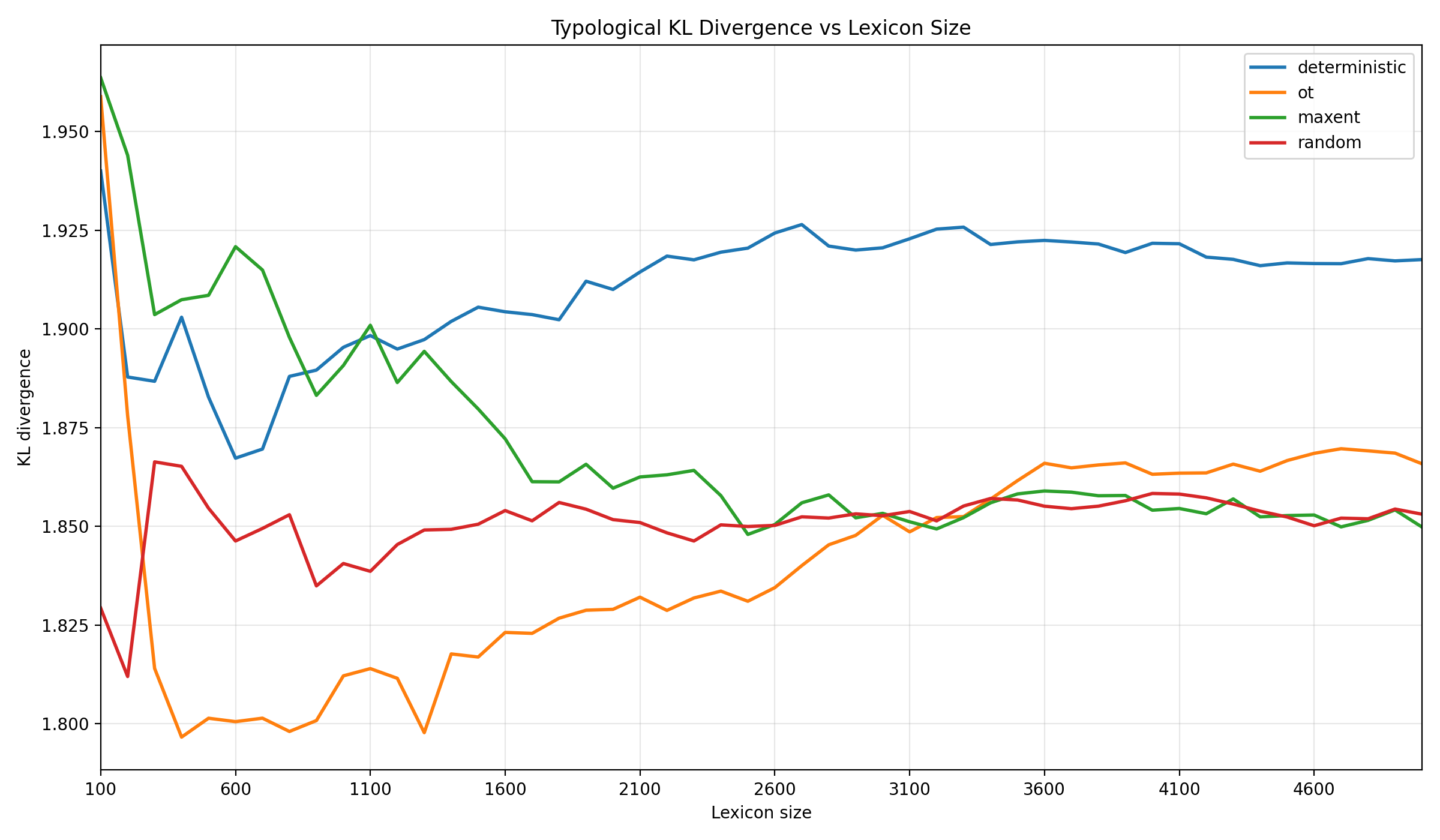}
    \caption{KL divergence $D_{\mathrm{KL}}(P \,\|\, Q)$ between the
    generated phoneme distribution and the PHOIBLE global reference,
    as a function of lexicon size. Lower values indicate greater
    typological realism.}
    \label{fig:kl}
\end{figure}
 
Figure~\ref{fig:kl} shows an unexpected dissociation from the
phonotactic results: the deterministic generator, which performed
second-best on well-formedness, exhibits the \emph{highest} KL
divergence at convergence (${\approx}1.915$), while MaxEnt converges
alongside the random baseline (${\approx}1.852$--$1.855$). This
dissociation arises from a fundamental difference between the two
evaluation axes. Phonotactic well-formedness rewards internal
consistency — the degree to which word forms obey the same structural
patterns — whereas typological realism rewards distributional breadth.
Hard-filtering in the deterministic generator enforces consistency by
suppressing entire syllable structures, but this categorical exclusion
systematically under-samples certain phoneme types, skewing the
generated distribution away from the cross-linguistic reference.
MaxEnt, by contrast, samples probabilistically over the full candidate
space, allowing rare phoneme contexts to surface occasionally and
incidentally recovering a distribution close to PHOIBLE's. That the
random baseline matches MaxEnt here is not a failure of MaxEnt but a
ceiling effect: once constraint enforcement is sufficiently gradient,
typological alignment is largely determined by the inventory
composition rather than the grammar.

\subsection{Cross-Grammar Compatibility}
\label{sec:results_cross}
 
\begin{figure}[t]
    \centering
    \begin{subfigure}{0.48\linewidth}
        \centering
        \includegraphics[width=\linewidth]{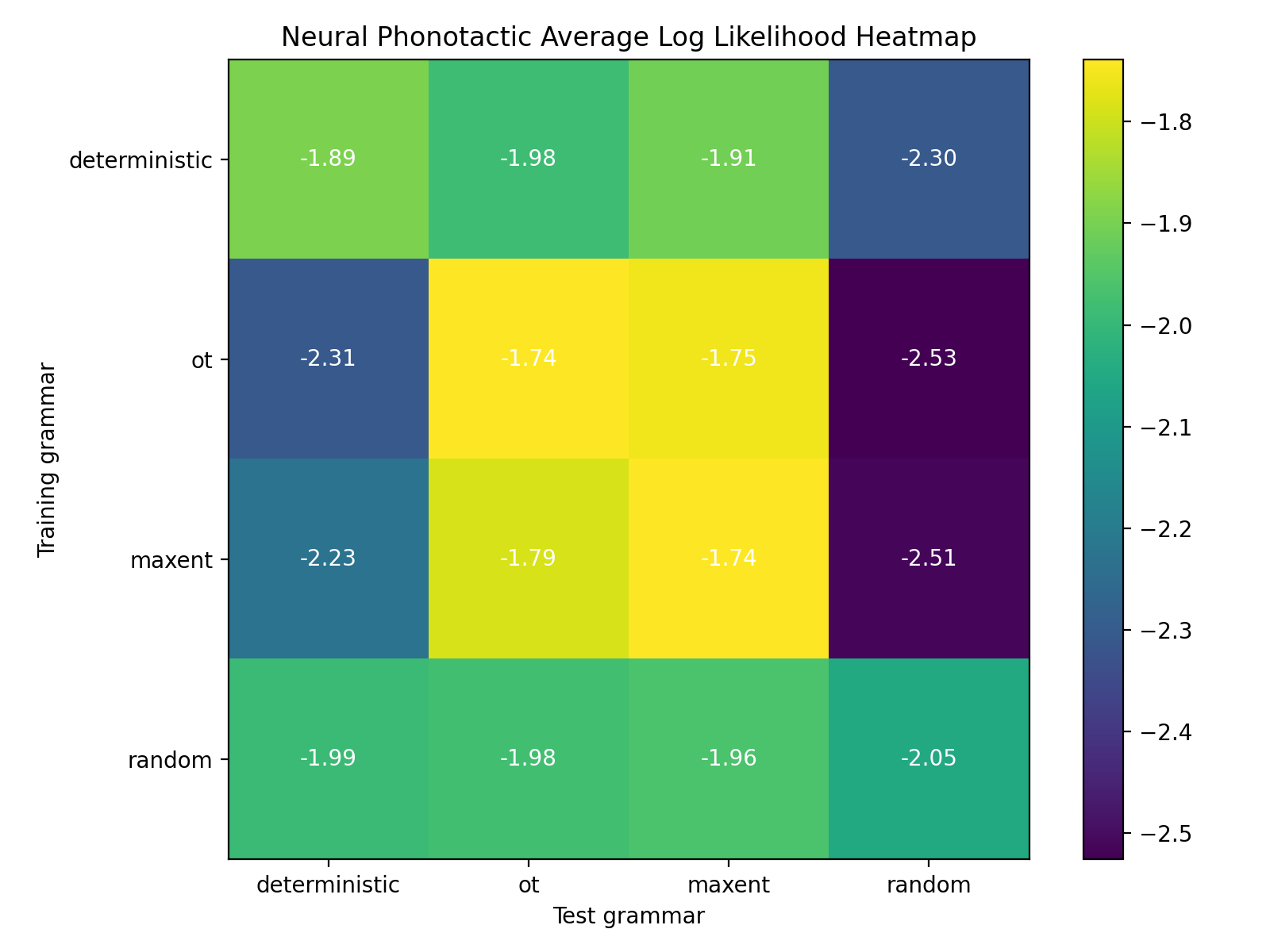}
        \caption{Average log-likelihood (higher is better).}
        \label{fig:cross_loglik}
    \end{subfigure}
    \hfill
    \begin{subfigure}{0.48\linewidth}
        \centering
        \includegraphics[width=\linewidth]{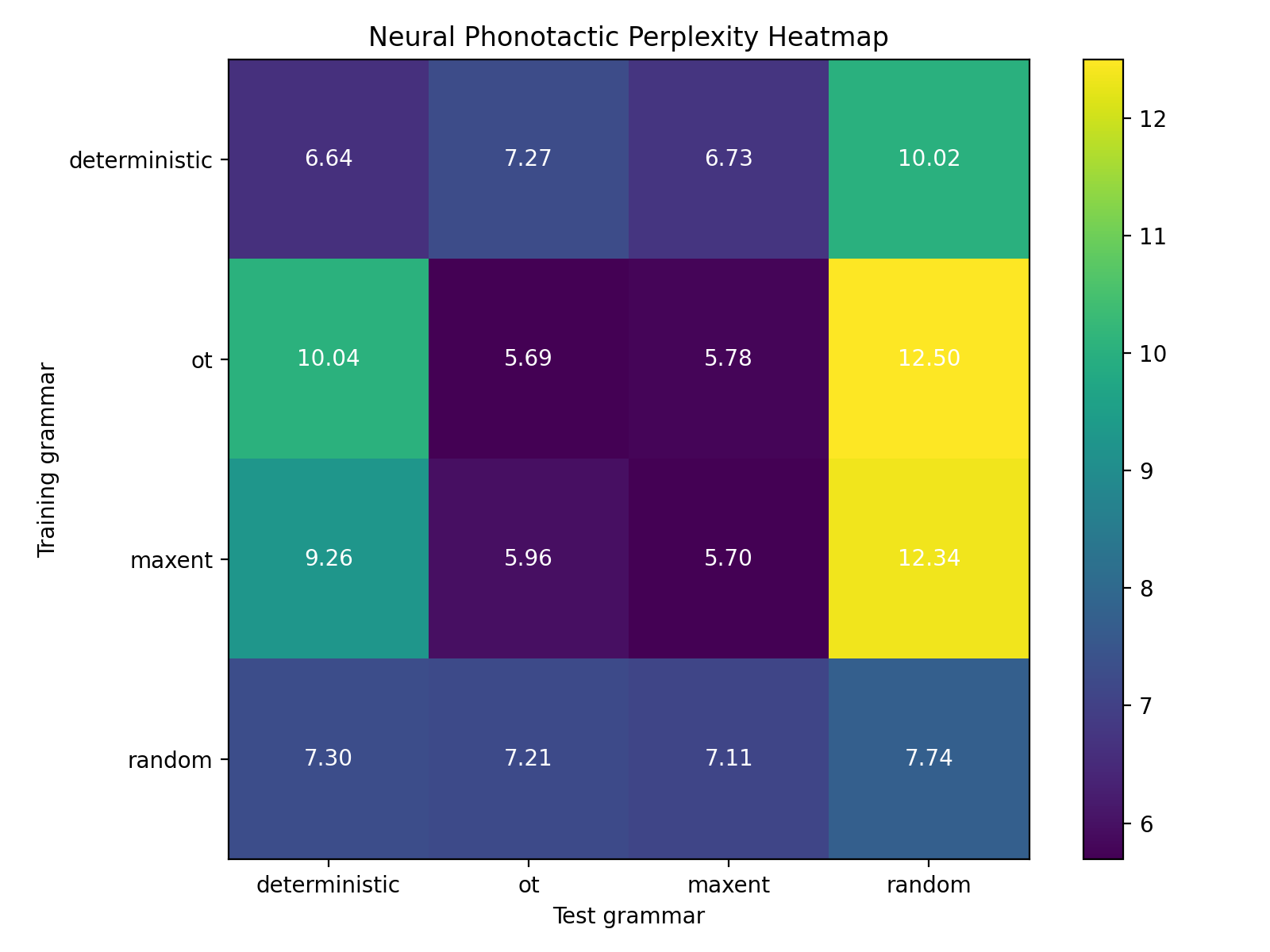}
        \caption{Perplexity (lower is better).}
        \label{fig:cross_perplexity}
    \end{subfigure}
    \caption{Cross-grammar evaluation matrices. Rows denote the
    training grammar; columns denote the test grammar. Diagonal
    entries reflect within-grammar fit; off-diagonal entries reveal
    distributional transferability across formalisms.}
    \label{fig:cross_heatmaps}
\end{figure}
 
\paragraph{OT and MaxEnt are distributionally equivalent.}
The OT--MaxEnt and MaxEnt--OT off-diagonal entries ($-1.75$ and
$-1.79$ log-likelihood; $5.78$ and $5.96$ perplexity) are nearly
indistinguishable from the respective diagonal entries ($-1.74$ and
$-1.74$; $5.69$ and $5.70$). A model trained on OT lexicons
explains MaxEnt lexicons almost as well as it explains its own —
and vice versa. This near-symmetry indicates that OT and MaxEnt
induce effectively the same phonotactic distribution, consistent
with the theoretical relationship between the two: MaxEnt is a
probabilistic relaxation of the same weighted constraint evaluation
that underlies OT.
 
\paragraph{Deterministic is a subset of the probabilistic space.}
The deterministic row shows moderate cross-grammar log-likelihoods
on OT and MaxEnt test sets ($-1.98$ and $-1.91$), indicating that
a model trained on deterministic lexicons partially transfers to
probabilistic ones. The converse is markedly asymmetric: OT and
MaxEnt models assign substantially lower likelihood to deterministic
test lexicons ($-2.31$ and $-2.23$). This directional asymmetry
implies that the deterministic grammar generates a restricted
subspace of the phonotactic patterns covered by probabilistic
grammars — its word forms are partially explained by probabilistic
models, but probabilistic word forms fall outside what the
deterministic model has learned.
 
\paragraph{Random lexicons are the universal outlier.}
All trained models assign their worst likelihoods to random test
lexicons, and the random-trained model transfers poorly to all
grammars. Crucially, the random model's diagonal entry ($-2.05$
perplexity $7.74$) is substantially worse than even the
deterministic diagonal, confirming that constraint enforcement —
even categorical — produces genuinely learnable structure absent
from unconstrained generation.

\newpage

\section*{Acknowledgments}
The authors acknowledge the use of AI tools such as ChatGPT, Claude, and Gemini for improving the presentation and grammar of this paper. All the results, analysis, and proposed techniques remain a concrete representation of the author's contributions. The authors take full responsibility for the contents in this paper.

\bibliography{colm2026_conference}
\bibliographystyle{colm2026_conference}

\appendix
\section{Typological Universals Evaluated in the Phoneme Sampler}
\label{app:universals}

The phoneme inventory sampler incorporates three categories of
typological tendencies, each validated over PHOIBLE inventories
\citep{moran2019phoible} prior to integration. Correlational
tendencies inform the probability of marked segment inclusion as
a function of inventory size; co-occurrence tendencies constrain
which feature combinations are permitted; and implicational
universals enforce directional dependencies between features,
either categorically ($P = 1.0$) or probabilistically where
near-universal violations are attested.

\paragraph{Correlational tendencies (Pearson $r$).}
Symmetric associations between inventory size and marked segment
presence, measured using Pearson correlation.
\begin{itemize}
    \item Inventory size $\leftrightarrow$ presence of clicks
    \item Consonant count $\leftrightarrow$ presence of ejectives
    \item Consonant count $\leftrightarrow$ presence of clicks
    \item Vowel count $\leftrightarrow$ presence of vowel length contrast
\end{itemize}

\paragraph{Co-occurrence tendencies (chi-square $\chi^2$).}
Symmetric associations between pairs of binary phonological
features, measured using chi-square tests of independence. Values
of $\chi^2 = 0$ reflect near-deterministic contingency structure
rather than independence.
\begin{itemize}
    \item Ejective consonants $\leftrightarrow$ uvular consonants
    \item Pharyngeal consonants $\leftrightarrow$ uvular consonants
    \item Nasal vowels $\leftrightarrow$ oral vowels
    \item Voiced obstruents $\leftrightarrow$ voiceless obstruents
    \item Fricatives $\leftrightarrow$ stops
    \item Front rounded vowels $\leftrightarrow$ front unrounded vowels
\end{itemize}

\paragraph{Implicational universals (conditional probability $P(Y \mid X)$).}
Directional dependencies of the form $X \Rightarrow Y$, measured
as conditional probabilities over inventories containing $X$.
Universals with $P = 1.0$ are enforced deterministically in the
sampler; the one near-universal ($P = 0.997$) is enforced
probabilistically.
\begin{itemize}
    \item Pharyngeal consonants $\Rightarrow$ uvular consonants
    \item Nasal vowels $\Rightarrow$ oral vowels
    \item Voiced obstruents $\Rightarrow$ voiceless obstruents
          \hfill{\small($P = 0.997$, near-universal)}
    \item Fricatives $\Rightarrow$ stops
    \item Front rounded vowels $\Rightarrow$ front unrounded vowels
\end{itemize}

\section{Code and Reproducibility}
\label{app:code}

All code, evaluation scripts, generated lexicons, and experimental
results reported in this paper are publicly available at:
\begin{center}
    \url{https://anonymous.4open.science/r/lexicon-generation/}
\end{center}
The repository includes the phoneme inventory sampler, all four
phonological grammar implementations (deterministic, OT, HG, and
MaxEnt), the semantic assignment module, and the phonotactic and
typological evaluation pipelines described in
Sections~\ref{sec:methodology} and~\ref{sec:evaluation}.

\end{document}